\documentclass[journal]{IEEEtran}

\makeatletter
    
    \newcommand{\Rmnum}[1]{\expandafter\@slowromancap\romannumeral #1@}
    \newcommand{\tabincell}[2]{\begin{tabular}{@{}#1@{}}#2\end{tabular}}

\newtheorem{proposition}{Proposition}
\newtheorem{lemma}{Lemma}
\newtheorem{definition}{Definition}
\newtheorem{remark}{Remark}

\usepackage{mathrsfs,amssymb,amsmath,booktabs,array,xcolor,url,cite,color,soul,multirow,multicol}
\usepackage{mathbbold,bm,bbm}
\usepackage{slashbox}
\usepackage[dvips]{graphicx}
\usepackage{textcomp}
\usepackage{ulem}
\usepackage{cases}

\ifCLASSINFOpdf
\else
\fi

\begin{document}
%
\title{{Enhancing Interpretability of Black-box Soft-margin SVM by Integrating Data-based Priors}}
%
%
%
%

\author{Shaohan~Chen,~Chuanhou~Gao,~\IEEEmembership{Senior Member,~IEEE,}~and~Ping~Zhang
\thanks{Manuscript received Aug. 15, 2018. This work was supported by
the National Natural Science Foundation of China under Grant No.
11671418 and 61611130124.}
\thanks{The authors are with the School of Mathematical Sciences, Zhejiang
University, Hangzhou 310027, China (e-mail: gaochou@zju.edu.cn).}}

\IEEEtitleabstractindextext{%
\begin{abstract}
The lack of interpretability often makes black-box models difficult
to be applied to many practical domains. For this reason, the
current work, from the black-box model input port, proposes
to incorporate data-based prior information into the black-box soft-margin SVM
model to enhance its interpretability. The concept and
incorporation mechanism of data-based prior information are
successively developed, based on which the interpretable or partly
interpretable SVM optimization model is designed and then solved
through handily rewriting the optimization problem as a nonlinear
quadratic programming problem. An algorithm for mining data-based linear
prior information from data set is also proposed, which generates a linear
expression with respect to two appropriate inputs identified from all inputs
of system. At last, the proposed interpretability enhancement strategy is applied to eight benchmark examples for effectiveness exhibition.
\end{abstract}

}

\maketitle

\IEEEdisplaynontitleabstractindextext

%
\IEEEpeerreviewmaketitle

\ifCLASSOPTIONcompsoc
\IEEEraisesectionheading{\section{Introduction}\label{sec:introduction}}
\else
\section{Introduction}
\label{sec:introduction}
\fi

\IEEEPARstart{D}{evelopment} of black-box modeling techniques, like
support vector machine (SVM), neural networks, etc., has shown
rather rapid in the past decades.
This sort of techniques, compared to white-box modeling methods
(also called mechanism-based modeling or first-principles modeling),
works without any need of knowing the internal structure or details
on variables interaction in systems considered, so they are
suited to describe extremely complex objectives, such as human brain
\cite{Khosrowabadi}, black hole \cite{Grumiller}, integrated
industrial processes \cite{Gao12} and so on. Essentially, black-box
modeling is an input-output data-based approach, and the model
precision mainly depends on data quality, model structure and
parameters identification algorithm. In order to develop
high-precision black-box models, it always needs reliable and
representative data, smart mathematical treatment and efficient
identification algorithms. All of these are challenging the
development of the black-box modeling techniques. Intuitively, it is
not always a good strategy to further develop advanced mathematical
methods for the improvement of the black-box models precision. Moreover, even if black-box models are accurate enough, a clear insight into the reasoning made by them is not available. Namely,
there is a severe lack of comprehensibility on the
operating principle of black-box models. The loss of interpretability
makes it impossible to explain the model outputs as comprehensive
knowledge, and neither to improve the model performance using known
knowledge about systems. Since explanation is one of the most important aspects that affects end users to accept models, the applications of black-box models in practical domains are restricted greatly. This is specially true for areas like credit risk analysis and medical diagnosis, where the definite causal for making a decision is desired and necessary. It thus becomes quite significant to investigate ways of enhancing interpretability of black-box models, which has been the goals of DARPA's currently ongoing Explainable AI project \cite{RSG16}. Of particular interest is the algorithm called to Local
Interpretable Model-Agnostic Explanations (LIME). LIME seeks to extract expressivity from black-box models.

\begin{figure}
\begin{center}
{
\includegraphics[width=3.3in]{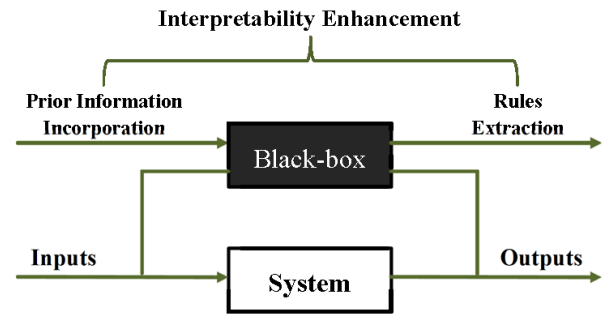}}
{\caption{Interpretability enhancement ways for the black-box model.}}\end{center}
\label{Black-box}
\end{figure}

There are two ways to make a black-box model more interpretable. One
way is from the input port of the black box to integrate prior
information into the model \cite{Lauer08,Hu11,Borges}, the other way
is from the output port to extract comprehensive rules from the
model \cite{Martens09,Chorowski,Huynh11}. A schematic diagram to
display these two interpretability enhancement ways is given in Fig. $1$. In this
paper, we will focus on the first strategy while the rules extracting
strategy may be found in the recent review given by Barakat et. al.
\cite{Barakat}. {The main reason is that the prior information is crucial for building models of the problem at hand. Its importance may be seen from no free lunch theorem\mbox{\cite{Wolpert}} that states all algorithms perform the same when averaged over different problems. Lauer et al.\mbox{\cite{LauerB08}} even pointed out that a model without prior knowledge is an ill-posed problem. The priors incorporation is thought as the unique means for a model to be extended into practice\mbox{\cite{Niyogi}} in the case that the data size is finite. More importantly, the inclusion of priors into the model can add both of interpretability and precision. Furthermore, unlike rule extraction only giving interpretability in knowledge reasoning, the prior information may refer to any aspect about the problem, like the structure, parameters, etc., so it can provide interpretability from every aspect\mbox{\cite{Hu07}}. }

Prior information is any known information on the problem
investigated beforehand. The forms of prior information are very
diverse, some about the model structure, some about the inherent
constraint, while some about the data example. For the successful
incorporation of prior information, it is necessary to sort out them
right. Broadly speaking, prior information is usually classified into two
main categories: knowledge about the estimated function, such as
smoothness \cite{Vapnik98}, symmetry \cite{Chen08}, monotonicity
\cite{Abu-Mostafa,Daniels}, boundary constraint/input domain
\cite{Lauer08,Mangasarian04,Mangasarian07,Mangasarian08}, etc., and
knowledge on the data \cite{LauerB08}, such as quality of the data,
including that if the data set is persistent exciting, if the data
set contains visible outliers, and if the data set is imbalanced
(mainly in the case of classification problem, a high proportion of
samples belongs to the same class). According to categories, the
prior information is embedded into the model using different modes.
The smoothness and symmetry may be used to define the model
structure \cite{Poggio,Aguirre}; the monotonicity and concavity may
be converted to the derivatives information of the estimated function, and are further incorporated in the form of the inequality constraints \cite{Lauer08} into the model; the
boundary constraint/input domain is embedded as equality or
inequality constraint \cite{Mangasarian04,Mangasarian07,Mangasarian08}. For the
information of non-persistent exciting data, Niyogi et. al. \cite{Niyogi} proposed to create virtual samples to enlarge the data set, while for the imbalanced data, the idea of weighting the
samples may trade off the data difference to weaken imbalance
\cite{Cristianini00}. In fact, it is not often
easy to recognize the categories or the incorporation methods of the
prior information explicitly. For examples, the boundary
constraint/input domain is related to both the estimated function
and the data, and the data knowledge is incorporated through affecting the estimated function.

{Another relatively distinct classification method may be based on the black-box models that need to be enhanced interpretability. Typical black-box models of concern include neural network and SVM. These two kinds of models have different structure, which leads to the incorporation means of priors very different. The former has a hierarchical structure with the incorporation of priors to modify the weights, bias, and/or minimize back-propagation errors. Towell et al.\mbox{\cite{Towell94}} mapped ``domain theories" in the form of propositional logic into network, and got stronger generalization ability. Daniels and his coworkers showed universal approximation capabilities of partial monotone\mbox{\cite{Minin10}} and monotone neural network\mbox{\cite{Daniels10,Velikova06}}, where partial monotonicity or monotonicity is incorporated by structure. Dugas et al.\mbox{\cite{Dugas09}} also confirmed that the generalization performance of neural network can be improved if the functional knowledge, like convexity and monotonicity, is incorporated. Hu et. al.\mbox{\cite{Hu07}} presented three
embedding modes for neural network from ``structural", ``algorithm" and ``data", and ranked them in a descending order with respect to transparency. Hu et al.\mbox{\cite{Hu16}} developed a general framework for deep neural network to incorporate and automatically optimize vast amount of fuzzy knowledge. As for the SVM model, it is of a constraints-optimization structure, so the prior information is usually incorporated either in the form of equality constraint\mbox{\cite{Poggio,Aguirre,Cristianini00}} or in the form of inequality constraint\mbox{\cite{Lauer08,Mangasarian04,Mangasarian07,Mangasarian08}}. More details may be found in the review paper\mbox{\cite{LauerB08}}.}

\begin{table*}
\caption{{Notations}} \label{NOTE} \centering
\begin{tabular}{l l}
\toprule
Symbol & Meaning \\
\midrule[0.06em]
     $F_1,F_2,F_3$ & Three functions related to the dual target margin, ``positive and negative class" prior information, respectively \\
     $\tilde{F}_2,\tilde{F}_3$ & Improved version of $F_2$ and $F_3$ through the Lagrangian multiplier $\boldsymbol{\alpha}$\\
     $G_1,G_2$ &  Two functions representing the terms related to the kernel function and the Lagrangian multipliers, respectively \\
     $\textbf{k}(\cdot,\cdot)$ & $N$-dimensional kernel function vector\\
     $l_1$,$l_2$,$l_3$ & The loss functions for incorrect classification, incorrect ``positive class" and ``negative class" prior information, respectively\\
     $N$ & The number of training samples\\
     $n$ & The dimension of data set\\
     $\mathbb{R}^n$ & The space of $n$-dimensional real vectors\\
     $\bar{\mathbb{R}}^n_+$ & The set of $n$-dimensional real vectors consisting of all nonnegative entries\\
     $\mathbb{R}^n_+$ & The set of $n$-dimensional positive real vectors\\
     $g^+,g^-$ & Two maps from $\Gamma^+$ and $\Gamma^-$ to $\mathbb{R}^{r^+}$ and $\mathbb{R}^{r^-}$, respectively, expressing ``positive and negative class" prior information \\
     $r^+,r^- $ & Two positive integers\\
     $t^+,t^-$ & The number of training samples contained in $\Gamma^+$ and $\Gamma^-$, respectively\\
     $\textbf{x}^-,\textbf{x}^-$ & Sampling input to render ``positive/negative class" prior information\\
     $\boldsymbol{\alpha},\boldsymbol{\beta},\boldsymbol{\gamma}$ & Lagrangian multiplier vectors\\
     $\widetilde{\boldsymbol{\beta}},\widetilde{\boldsymbol{\gamma}}$ & Improved version of Lagrangian multiplier vectors of $\boldsymbol{\beta}$ and $\boldsymbol{\gamma}$, respectively\\
     $\Gamma^+,\Gamma^-$ & The subset of $\mathbb{R}^n$ containing sampling input $\textbf{x}^-,\textbf{x}^-$, respectively \\
     $\mathcal{L}(\boldsymbol{\alpha})$& The dual target margin \\
     $\mathscr{L}_{ij}^+,\mathscr{L}_{ij}^-$ & Labeling linear ``positive and negative class" prior information in the plane $X^{(i)}OX^{(j)}$ composed of two features $x^{(i)}$ and $x^{(j)}$\\
     $\boldsymbol{\zeta},\boldsymbol{\varsigma}$ & Slack variable vectors to measure incorrectness of the mined ``positive and negative class" prior information, respectively\\
     $\theta$ & Kernel slack variable\\
     $\lambda_1,\lambda_2,\lambda_3$ & Counterbalance constants\\
     $\boldsymbol{v}_+,\boldsymbol{v}_-$ & $r^+$ and $r^-$ dimensional nonnegative real vectors, respectively\\
     $\Phi$ & Feature map from $\mathbb{R}^n$ to high dimensional feature space\\
     $\mathbbold{0}_n$,$\mathbbm{1}_n$ & $n$-dimensional vector with all entries equal to $0$ and $1$, respectively\\
     s.t. & The abbreviation of ``subject to"\\
\bottomrule
\end{tabular}
\end{table*}

This work continues to enrich the methods of integrating the prior
information into the black-box model. {We take the soft-margin SVM\mbox{\cite{Scholkopf,Xu13}} as an example of black-box models to be incorporated with the priors for interpretability enhancement. The main contributions include the development of an optimization-based linear priors mining algorithm from data, and the construction plus solving of the partly interpretable soft-margin SVM model. In the model construction, we fully consider the correctness of the mined linear priors, and design the interpretable model as the balance of maximizing the margin and minimizing the errors of the priors. In the model solving process, we rewrite the interpretable model to have the same structure as the pure black-box soft margin SVM so that the common SVM software package\mbox{\cite{Chang01}} can be used directly. The mined linear priors are embedding into the black-box model in the form of inequality constraints\mbox{\cite{Mangasarian04,Mangasarian07,Mangasarian08}}, which add interpretability of the black-box model by affecting the model structure and further the solving algorithm. Additionally, the model precision is also enhanced from incorporating the mined linear priors, since some ``important" samplings are expected to be classify right. We use eight benchmark examples to evaluate the performance of the proposed interpretability enhancement method. }

To ensure the black-box soft-margin SVM model working more
practically, the following assumptions are made on pursuing the
current research:

\textit{A1. In all probability, there are non-separable or
mislabeled samples when classification is executed on real-world
data;}

\textit{A2. The prior information acquired from real-world data
allows to be not exact as the true one.}

Note that these two assumptions are so general in practice that this
work can serve for extensive applications. The rest of this paper is
organized as follows. Section II introduces preliminary on SVM and its soft-margin version. Then, formulation of data-based prior information is proposed in Section III. Next, Section IV presents a
interpretability enhancement pattern of the soft-margin SVM
through incorporating data-based prior information. This is followed
by some numerical experiments exhibition in Section V, including eight benchmark
examples and two real blast furnace examples. Finally, Section VI
concludes this paper.

Throughout the paper, the bold typeface denotes a vector or a matrix while the normal typeface stands for a scalar. The transpose of a vector or a matrix is denoted by the superscript ``$\top$",
while the superscript apostrophe denotes the derivative of a function with respect to its argument. For any two vectors $\textbf{z}_a=(z_{a_1},\cdots,z_{a_n})^\top$ and $\textbf{z}_b=(z_{b_1},\cdots,z_{b_n})^\top\in\mathbb{R}^n$ the inequality $\textbf{z}_a\geq \textbf{z}_b$ means $z_{a_i}\geq z_{b_i},~\forall i$. More notations may be found in Table $1$.

\section{Soft-margin Support Vector Machine}
Consider a binary classification
problem. Let $\textbf{x}\in\mathbb{R}^n$, $y\in\{+1,-1\}$ represent the input pattern and output pattern, respectively. The $\nu$-SVM \cite{Scholkopf} aims at solving the optimization problem
\begin{eqnarray}\label{MSVM2}
&\displaystyle\ & \min_{\textbf{w},b,\rho\geq 0,\xi_i\geq 0}\ \ -\nu\rho+\frac{1}{2}\textbf{w}^\top\textbf{w}+\frac{1}{N}\sum_{i=1}^{N}\xi_i, \notag\\
&\text{s.t.} \ \ &y_i(\textbf{w}^\top\Phi(\textbf{x}_i)+b)\geq
\rho-\xi_i,\forall i,
\end{eqnarray}
where $\textbf{w}$ is the normal vector in an imaginary
high-dimensional feature space, $N\in\mathbb{R}$ is the number of samplings, $b\in\mathbb{R}$ is the offset,
$\Phi(\cdot)$ is a high-dimensional feature project from $\mathbb{R}^n$ to the feature space, and $\nu\in[0,1]$ represents the upper bound on the fraction of training errors or the lower bound on the fraction of support vectors. If
$\textbf{w}^\top\Phi(\textbf{x}_i)+b\geq \rho-\xi_i$ for a $\textbf{x}_i$ then this $\textbf{x}_i$ belongs to the class of $y_i=+1$ while if
$\textbf{w}^\top\Phi(\textbf{x}_i)+b\leq -\rho+\xi_i$ then $\textbf{x}_i$ belongs to the
class of $y_i=-1$. Here, $\xi_i~(i=1,\cdots,N)$ are non-negative slack variables
to measure the degree of misclassification of the data $\textbf{x}_i$. By utilizing Lagrangian multipliers $\boldsymbol{\alpha}=(\alpha_1,\cdots,\alpha_N)^\top\in\bar{\mathbb{R}}^N_+$ and further making dual transformation, the optimization problem of Eq. (\ref{MSVM2}) can be written as
\begin{eqnarray}\label{MSVM3}
&\displaystyle &\max_{\boldsymbol{\alpha}}\ \
\mathcal{L}(\boldsymbol{\alpha}),\notag\\
&\text{s.t.} \ \ \ \
&\boldsymbol{\alpha}^\top\textbf{y}=0,\mathbbold{0}_N\leq
\boldsymbol{\alpha}\leq \frac{1}{N}\mathbbm{1}_N, \boldsymbol{\alpha}^\top\mathbbm{1}_N\geq\nu,
\end{eqnarray}
where
$\mathcal{L}(\boldsymbol{\alpha})=-\frac{1}{2}\sum_{i,j=1}^N\alpha_i\alpha_jy_iy_jk(\textbf{x}_i,\textbf{x}_j)$, and $k(\textbf{x}_i,\textbf{x}_j)$ is the well-known kernel function \cite{Cristianini00} defined by
\begin{eqnarray}\label{kernel}
&\displaystyle\
&k(\textbf{x}_i,\textbf{x}_j)=\Phi(\textbf{x}_i)^\top\Phi(\textbf{x}_j).
\end{eqnarray}
The decision function for $\nu$-SVM takes the form of
\begin{eqnarray}\label{decision2}
&\displaystyle\
&f(\textbf{x})=\text{sign}\Big(\sum_{\alpha_i\textgreater
0}^{i=1\sim N}\alpha_iy_ik(\textbf{x},\textbf{x}_i)+b\Big).
\end{eqnarray}

The optimization problem of Eq. (\ref{MSVM3}) may be further relaxed by introducing the nonnegative
kernel slack variable $\theta$ \cite{Xu13}, defined
by the difference of the target margin $\tau$ and the above dual
target $\mathcal{L}(\boldsymbol{\alpha})$, i.e.,
\begin{eqnarray}\label{slack-kernel}
&\displaystyle &\theta=\tau-\mathcal{L}(\boldsymbol{\alpha}).
\end{eqnarray}
Making a balance between the maximum margin and the minimum error penalty can create a new optimization problem
\begin{eqnarray}\label{MSVM4}
&\displaystyle &\min_{\tau,\boldsymbol{\alpha}\in \mathcal{A}, \theta\geq 0}\ \ -\tau+\lambda_1l_1(\theta),\notag\\
&\text{s.t.} \ \ \ \ &\mathcal{L}(\boldsymbol{\alpha})\geq
\tau-\theta,
\end{eqnarray}
where
$\mathcal{A}=\{\boldsymbol{\alpha}|\boldsymbol{\alpha}^\top\textbf{y}=0,\mathbbold{0}_N\leq
\boldsymbol{\alpha}\leq \frac{1}{N}\mathbbm{1}_N,
\boldsymbol{\alpha}^\top\mathbbm{1}_N\geq\nu\}$, the parameter
$\lambda_1\in \mathbb{R}_+$ acts as a counterbalance, and
$l_1(\cdot)$ represents any loss function. Note that the introduction of $\theta$ means that the margin of classification changes from the hard one $\tau$ to the soft one $\tau-\theta$. We thus refer to Eq. (\ref{MSVM4}) as soft-margin SVM in the following. However, the solutions will keep unchanged for these two optimization problems.

\begin{proposition}
Soft-margin SVM of Eq.
(\ref{MSVM4}) has the same solutions as $\nu$-SVM of Eq. (\ref{MSVM3}).
\end{proposition}

\begin{IEEEproof}
This result is a special case of \textit{Proposition 2} in the paper \cite{Xu13}. See the detailed proof there.
\end{IEEEproof}

\section{Data-based Prior Information}
\subsection{Prior Information}
Prior information incorporated into black-box models will
add a high degree of interpretability. Moreover, if the size of data is limited, this incorporation is thought as the sole means to improve the generalization performance of black-box models \cite{Niyogi}. Here, prior information is defined as follows.

\begin{definition}
\cite{Hu11}. Prior information refers
to any known information about or related to the concerning objects,
such as data, knowledge, specifications, etc.
\end{definition}

In this work, attention is mainly focused on the prior information related to data collected. However, it is not statistics of data, or characteristics that can be directly observed from data, such as imbalance, but
logical implications acquired by data-mining techniques. These logical implications are quite like the
nonlinear prior knowledge proposed by Mangasarian et. al.
\cite{Mangasarian04,Mangasarian07,Mangasarian08}. In the following,
we will shortly introduce the formulation of that kind of prior
knowledge \cite{Mangasarian08} within the framework of SVM black-box model.

Let the training examples be
$\{(\textbf{x}_i,y_i)\}^N_{i=1}$, $\Gamma^+,~\Gamma^- \subseteq
\mathbb{R}^n$ and $g^+:\Gamma^+ \to \mathbb{R}^{r^+}$, $g^-:\Gamma^-
\to \mathbb{R}^{r^-}$, where $r^+ $ and $r^- $ are two positive integers, then the nonlinear prior information is
expressed as
\begin{equation}\label{pi1}
g^+(\textbf{x}^+)\leq \mathbbold{0}_{r^+} \Longrightarrow \sum_{i=1}^
N\alpha_iy_ik(\textbf{x}^+,\textbf{x}_i)+b \geq b^*,~~\forall~
\textbf{x}^+\in \Gamma^+
\end{equation}
to identify positive class points $y=+1$, and
\begin{equation}\label{pi2}
g^-(\textbf{x}^-)\leq \mathbbold{0}_{r^-} \Longrightarrow \sum_{i=1}^
N\alpha_iy_ik(\textbf{x}^-,\textbf{x}_i)+b \leq -b^*,\forall~
\textbf{x}^-\in \Gamma^-
\end{equation}
to classify negative class points $y=-1$. Here, $b^*\in \bar{\mathbb{R}}_+$ is often set $0$
or $1$ in practice. Mathematically, the above prior information, as an example of
the ``positive class" prior information of Eq. (\ref{pi1}), means that
\begin{equation}\label{pi1a}\left.
\begin{array}{rr}
g^+(\textbf{x}^+)\leq \mathbbold{0}_{r^+}, \\
\textbf{z}^\top\textbf{k}(\textbf{x}^+)+b \textless b^*
\end{array}
\right\} ~~~ \text{has no solution},
\end{equation}
where $\textbf{k}(\textbf{x}^+)=(k(\textbf{x}^+,\textbf{x}_1),\cdots,k(\textbf{x}^+,\textbf{x}_N))^\top$ and $\textbf{z}=(\alpha_1y_1,\cdots,\alpha_Ny_N)^\top$.
Moreover, if $\Gamma^+$ is a convex subset of
$\mathbbm{R}^n$, and $g^+(\textbf{x}^+)$ and
$\textbf{k}(\textbf{x}^+)$ are convex on $\Gamma^+$, the following
result holds.

\begin{lemma}
\cite{Mangasarian08}. The prior information
expressed as Eq. (\ref{pi1}) or Eq. (\ref{pi1a}) is equivalent to
the result that there exists $\boldsymbol{v}_+\in
\bar{\mathbb{R}}^{r^+}_+$ such
that
\end{lemma}
\begin{equation}\label{pi1b}
\textbf{z}^\top\textbf{k}(\textbf{x}^+)+b-b^*+\boldsymbol{v}_+^\top g^+(\textbf{x}^+)\geq
0.
\end{equation}

For the ``negative class" prior information of Eq. (\ref{pi2}),
there is also the corresponding parallel result.

\begin{lemma} \cite{Mangasarian08}. The prior information
expressed as Eq. (\ref{pi2}) is equivalent to the result that
there exists $\boldsymbol{v}_-\in \bar{\mathbb{R}}^{r^-}_-$ such that
\end{lemma}
\begin{equation}\label{pi2a}
-\textbf{z}^\top\textbf{k}(\textbf{x}^-)-b-b^*+\boldsymbol{v}_-^\top g^-(\textbf{x}^-)\geq
0.
\end{equation}

The work of Mangasarian et. al.
\cite{Mangasarian04,Mangasarian07,Mangasarian08} successfully
converted the prior information in the form of logical implication
into the corresponding nonlinear inequality constraint. Obviously, the
latter is more easily to be incorporated into black-box models. It has been shown in some numerical experiments that this incorporation can improve greatly precision of SVM models \cite{Mangasarian04,Mangasarian07,Mangasarian08}.

\subsection{Concept of Data-based Prior Information}
To incorporate prior information into black-box models, it is necessary to acquire it firstly. A feasible solution to tackle this issue is to mine it from data. As said in the assumption A2, there is a deviation between the mined prior information and the true one in all probability. The acquirement of prior information is thus modeled by minimizing the loss of its incorrectness.

We also consider a binary classification problem for the samples $\{(\textbf{x}_i,y_i)\}^N_{i=1}$. Let $h(\cdot)$ represent a classifier rendered by any black-box model, in the case of SVM model which takes the form of $h(\cdot)=f(\cdot)=\sum_{i=1}^N\alpha_iy_ik(\cdot,\textbf{x}_i)+b$. The classification occurs according to
\begin{equation}\label{decisionFunction}
y=\left\{
\begin{array}{ll}
+1,~~h(\textbf{x})\geq b^*,\\
-1,~~h(\textbf{x})\leq -b^*.
\end{array}
\right.
\end{equation}
Let $g^+(\cdot)\leq \mathbbold{0}_{r^+}$ and $g^-(\cdot)\leq \mathbbold{0}_{r^-}$ be the positive and negative class functions that need to be mined, respectively, then the positive class and negative class prior information are expressed as
\begin{equation}\label{pi3}
g^+(\textbf{x}^+)\leq \mathbbold{0}_{r^+} \Longrightarrow h(\textbf{x}^+) \geq b^*,~~\forall~
\textbf{x}^+\in \Gamma^+,
\end{equation}
and
\begin{equation}\label{pi2}
g^-(\textbf{x}^-)\leq \mathbbold{0}_{r^-} \Longrightarrow h(\textbf{x}^-) \leq -b^*,~~\forall~
\textbf{x}^-\in \Gamma^-.
\end{equation}
From the assumption A2, these two pieces of knowledge may be not true. We thus introduce nonnegative slack variables $\zeta$ and $\varsigma$ to measure their incorrectness, respectively. Based on \textbf{Lemmas} $1$ and $2$, they are defined by
\begin{equation}
\zeta=b^*-[
h(\textbf{x}^+)+\boldsymbol{v}_+^\top g^+(\textbf{x}^+)],~~\forall~
\textbf{x}^+\in \Gamma^+,
\end{equation}
and
\begin{equation}
\varsigma=b^*-[\boldsymbol{v}_-^\top g^-(\textbf{x}^-)-h(\textbf{x}^-)],~~\forall~
\textbf{x}^-\in \Gamma^-.
\end{equation}
The loss induced by these two error variables may be evaluated by any loss function, similar to penalizing $\theta$ in Eq. (\ref{MSVM4}) for soft-margin SVM \cite{Xu13}. Denote the loss function by $l_2(\cdot)$ and $l_3(\cdot)$ for $\zeta$ and $\varsigma$, respectively, then the penalty to the incorrectness
may create the following optimization problem
\begin{eqnarray}\label{dpi}
&\displaystyle \ \ &\min_{\zeta, \varsigma\geq 0}\ \lambda_2  l_2(\zeta)+\lambda_3  l_3(\varsigma), \\
&\text{s.t.}\ \ &h(\textbf{x}^+)-b^*+\boldsymbol{v}_+^\top g^+(\textbf{x}^+)+\zeta \geq 0,  ~~\forall~
\textbf{x}^+\in \Gamma^+,\nonumber\\
&\ \ \ \ \ \  &h(\textbf{x}^-)
+b^*-\boldsymbol{v}_-^\top g^-(\textbf{x}^-)-\varsigma \leq
0,  ~~\forall~
\textbf{x}^-\in \Gamma^-,\nonumber
\end{eqnarray}
where $\lambda_2,~\lambda_3\in \mathbb{R}_+$ are constants, called counterbalance, like $\lambda_1$ in Eq. (\ref{MSVM4}). The solution may suggest two pieces of prior information: 1) If $g^+(\textbf{x}^+)\leq \mathbbold{0}_{r^+}$, then $\textbf{x}^+$ has the class label $y=+1$; 2) If $g^-(\textbf{x}^-)\leq \mathbbold{0}_{r^-}$, then $\textbf{x}^-$ has the class label $y=-1$. Thus, we have the following result.

\begin{proposition}
For any system with input-output pairs \{$(\textbf{x}_i,y_i)$\}$_{i=1}^N$, where $\textbf{x}_i\in\mathbb{R}^n$ and $y_i\in \{+1,-1\}$, assume $h(\textbf{x})$ defined by Eq. (\ref{decisionFunction}) to be a decision function for addressing binary classification problem of this system. Then the prior information: 1) $g^+(\textbf{x}^+)\leq \mathbbold{0}_{r^+}\Longrightarrow y=+1$ and 2) $g^-(\textbf{x}^-)\leq \mathbbold{0}_{r^-}\Longrightarrow y=-1$ can be corrected furthest by solving the optimization problem of Eq. (\ref{dpi}).
\end{proposition}

\begin{IEEEproof} The result is straightforward from \textbf{Lemmas} $1$ and $2$.
\end{IEEEproof}

The incorrectness of the prior information mined through data mining techniques is fully considered in the above optimization problem, so we define it by data-based prior information.

\begin{definition}
If prior information in the form of logical implication is acquired by any data-mining technique, and the incorrectness is minimized through Eq. (\ref{dpi}), then the acquired prior information is called data-based prior information.
\end{definition}

\begin{remark}
The modification of data-based prior information depends strongly on the black-box model for which it serves. In practical operation, this modification process and the black-box model training should be implemented synchronously. A naive idea for this purpose is to merge their object functions and constraints, respectively, to form one optimization problem, which simultaneously finishes the incorporation of prior information into the black-box model.
\end{remark}

{It should be noted that if the data-based priors deviate from the true ones far, then the synchronous optimization on the objective function of black-box model and the incorrectness of data-based priors will destroy the precision of the black-box model. At this point, it is difficult to tune the regularization parameters in the integrated model to enhance both of the interpretability and precision. This may be also suggested by allowing non-separable or mislabelled samples and non-true priors in assumptions \textit{A1} and \textit{A2}, respectively. However, from the viewpoint of advancing practical applications, the interpretability enhancement is a little more urgent than high precision for black-box models, so the synchronous optimization is still a good strategy even if it may lead to slight precision reduction. This also constitutes our main motivation to integrate data-based priors into the black-box model to enhance
interpretability in the current work. Naturally, if the data-based priors are true, the synchronous optimization strategy is potential to result in the enhancement of both of interpretability and precision of black-box models. }

\section{Incorporation of Data-based Prior Information into Soft-margin SVM}
In this section, data-based prior information is incorporated into
soft-margin SVM for the purposes of increasing its interpretability and also maintaining high precision. The classifier in Eq. (\ref{dpi}) thus takes the form of $h(\cdot)=f(\cdot)$.

For soft-margin SVM model of Eq. (\ref{MSVM4}), it requests to minimize the margin error while for data-based prior information model of Eq. (\ref{dpi}) minimizing the priors incorrectness. Their incorporation is naturally made through minimizing the sum of two objective functions, as said in \textit{Remark 1}, i.e., making a trade-off between a large margin and small error penalties, which creates the following optimization problem
\begin{eqnarray}\label{Trans0}
&\displaystyle \ \ &\min_{\tau, \boldsymbol{\alpha}\in \mathcal{A},\theta\geq 0, \zeta\geq 0, \varsigma\geq 0}\ -\tau+\boldsymbol{\lambda}^\top\boldsymbol{l}, \\
&\text{s.t.} \ \ \ \ &\mathcal{L}(\boldsymbol{\alpha})\geq
\tau-\theta, \nonumber\\
&\ \ \ \ \ \  &f(\textbf{x}^+)-b^*+\boldsymbol{v}_+^\top g^+(\textbf{x}^+)+\zeta \geq 0,  ~~\forall~
\textbf{x}^+\in \Gamma^+,  \nonumber\\
&\ \ \ \ \ \  &f(\textbf{x}^-)+b^*-\boldsymbol{v}_-^\top g^-(\textbf{x}^-)-\varsigma \leq
0,  ~~\forall~
\textbf{x}^-\in \Gamma^-, \nonumber
\end{eqnarray}
where $\boldsymbol{\lambda}=(\lambda_1,\lambda_2,\lambda_3)^\top$ and $\boldsymbol{l}=(
l_1(\theta),l_2(\zeta),l_3(\varsigma))^\top.$

\begin{remark}
Although the data-based priors, i.e., the last two constraints in Eq. (\ref{Trans0}), are mined from the data samples $\{(\textbf{x}_i,y_i)\}^N_{i=1}$, which means these constraints naturally hold for some of the data points, they have to hold for all feasible (infinitely many) points in region of the feature space defined by the prior knowledge constraints. Namely, the prior knowledge constraints in Eq. (\ref{Trans0}) are semi-infinite \cite{GL98} in the variables $\textbf{x}^+$ and $\textbf{x}^-$, and the optimization problem of Eq. (\ref{Trans0}) is a semi-infinite optimization problem.
\end{remark}

For a semi-infinite optimization problem, if the SVM model and the prior knowledge constraints are linear \cite{FMS03}, the problem is easy to solve since the semi-infinite variables may be removed by applying theorems of the alternative \cite{Man94}. However, when any of them is nonlinear, this leads to a semi-infinite nonlinear optimization problem, which is very difficult to solve as it is NP-Hard. To circumvent this issue, Mangasarian and Wild \cite{Mangasarian08} discretize the semi-infinite variables into a uniform mesh, which may convert a semi-infinite program \cite{GL98} with an infinite number of constraints into a finite mathematical program. Note that the soft margin SVM is nonlinear and we also think the data-based priors to be nonlinear so that they are easy to be generalized. Therefore, Eq. (\ref{Trans0}) is a semi-infinite nonlinear optimization problem. We thus adopt the same discretizing technique as presented in \cite{Mangasarian08} to convert it into a finite mathematical program, i.e.,
\begin{eqnarray}\label{Trans1}
&\displaystyle \ \ &\min_{\tau, \boldsymbol{\alpha}\in \mathcal{A},\theta\geq 0, \boldsymbol{\zeta}\geq \mathbbold{0}_{t^+}, \boldsymbol{\varsigma}\geq \mathbbold{0}_{t^-}}\ -\tau+\boldsymbol{\lambda}^\top\boldsymbol{l}, \\
&\text{s.t.}  &\mathcal{L}(\boldsymbol{\alpha})\geq
\tau-\theta, \nonumber\\
&\ \ \ \ \ \  &f(\textbf{x}^+_j)-b^*+\boldsymbol{v}_+^\top g^+(\textbf{x}^+_j)+\zeta_j \geq 0, j=1,\cdots,q^+,  \nonumber\\
&\ \ \ \ \ \  &f(\textbf{x}^-_h)
+b^*-\boldsymbol{v}_-^\top g^-(\textbf{x}^-_h)-\varsigma_h \leq
0, h=1,\cdots,q^-, \nonumber
\end{eqnarray}
where $q^+$ and $q^-$ are the numbers of finite meshes of points for positive class prior and negative class prior, respectively, and $$\boldsymbol{l}=\Big(
l_1(\theta),\sum_{j=1}^{q^+} l_2(\zeta_j),\sum_{h=1}^{q^-}
l_3(\varsigma_h)\Big)^\top.$$
This model will act as the benchmark model that can enhance the interpretability but without loss of precision of the black-box soft-margin SVM. We refer to it and its equivalent models as partly interpretable soft-margin SVM, abbreviated to \textit{pTsm}-SVM.

A common strategy to solve the above optimization problem is to
convert it into the corresponding dual form, then we have the following proposition.

\begin{proposition}
The solution of pTsm-SVM in Eq. (\ref{Trans1})
is the same as that of the following optimization problem
\begin{equation}\label{Trans2}
\min_{\boldsymbol{\alpha}}\max_{
\boldsymbol{\beta}, \boldsymbol{\gamma}}~~ F_1(
\boldsymbol{\alpha})+F_2(
\boldsymbol{\alpha},\boldsymbol{\beta})+F_3(
\boldsymbol{\alpha},\boldsymbol{\gamma})+\boldsymbol{\lambda}^\top\boldsymbol{\ell},
\end{equation}
where
\begin{equation}\label{F1F2F3}\left\{
\begin{array}{ll}
F_1(\boldsymbol{\alpha})=-\mathcal{L}(\boldsymbol{\alpha}), \\
F_2(\boldsymbol{\alpha},\boldsymbol{\beta})=-
\sum_{j=1}^{q^+}\beta_j[f(\textbf{x}^+_j)-b^*+\boldsymbol{v}_+^\top g^+(\textbf{x}^+_j)],\\
F_3(\boldsymbol{\alpha},\boldsymbol{\gamma})=
\sum_{h=1}^{q^-}\gamma_h[f(\textbf{x}^-_h)
+b^*-\boldsymbol{v}_-^\top g^-(\textbf{x}^-_h)],
\end{array}
\right.
\end{equation}
and
\begin{equation*}
\boldsymbol{\ell}=\boldsymbol{l}-\Big(
l_1'(\theta)\theta,\sum_{j=1}^{q^+}
l_2'(\zeta_j)\zeta_j,\sum_{h=1}^{q^-}
l_3'(\varsigma_h)\varsigma_h\Big)^\top
\end{equation*}
with $l_1'(\cdot),l_2'(\cdot),l_3'(\cdot)$ representing the corresponding derivatives with respect to their respective argument. The decision variables
$\boldsymbol{\beta}=(\beta_1,\cdots,\beta_{N^+})^\top\geq
\mathbbold{0}_{N^+}$ and
$\boldsymbol{\gamma}=(\gamma_1,\cdots,\gamma_{N^-})^\top\geq
\mathbbold{0}_{N^-}$ are the Lagrangian multipliers.
\end{proposition}

\begin{IEEEproof}
The conversion from Eq. (\ref{Trans1}) to
Eq. (\ref{Trans2}) is easily realized by constructing the Lagrangian
function of the former, and then utilizing the corresponding KKT
conditions.
\end{IEEEproof}

\begin{remark}
The optimization model of Eq.
(\ref{Trans2}) will degenerate to be
\begin{equation}\label{Trans3}
\min_{\boldsymbol{\alpha}}\max_{
\boldsymbol{\beta}, \boldsymbol{\gamma}}~~ F_1(
\boldsymbol{\alpha})+F_2(
\boldsymbol{\alpha},\boldsymbol{\beta})+F_3(
\boldsymbol{\alpha},\boldsymbol{\gamma})
\end{equation}
if all the loss functions are selected as the hinge loss, i.e.,
$l(\cdot)=\text{max}~(0,\cdot)$ \cite{Xu13}. In this case, the decision
variables satisfy
$\mathbbold{0}_{q^+}\leq \boldsymbol{\beta}\leq
\lambda_2\mathbbm{1}_{q^+}$, $\mathbbold{0}_{q^-}\leq
\boldsymbol{\gamma}\leq \lambda_3\mathbbm{1}_{q^-}$ and
$\boldsymbol{\alpha}\in \mathcal{A}$.
\end{remark}

\begin{remark}
For the degenerated \textit{pTsm}-SVM in Eq. (\ref{Trans3}), the objective consists of three functions, $F_1(\boldsymbol{\alpha}),~F_2(\boldsymbol{\alpha},\boldsymbol{\beta})$ and $F_3(
\boldsymbol{\alpha},\boldsymbol{\gamma})$, which respectively measure
the contributions of pure SVM, ``positive class" and ``negative
class" prior information to the optimization objective. The first term represents the black-box part of model while the latter two terms express the white-box part of model. The ratio of ``black to white" in the model can be controlled through setting feasible domains of the decision variables.
\end{remark}

\begin{remark}
Besides enhancing interpretability, the priors incorporation into black-box soft-margin SVM also have potential to improve the model performance if they are true. On the one hand, these additional extra constraints can reduce the feasible domain of soft-margin SVM greatly, which provides larger opportunity to find global solutions; on the other hand, the acquired priors mean that some ``important" samplings are singled out from all training points, and these ``important" samplings are expected to be classified right with the trained model. Hence, these samplings will have a stronger effect on identifying model parameters.
\end{remark}

A further look at the expressions of $F_1(\boldsymbol{\alpha}),~F_2(\boldsymbol{\alpha},\boldsymbol{\beta})$ and $F_3(\boldsymbol{\alpha},\boldsymbol{\gamma})$ in Eq. (\ref{F1F2F3})
reveals that $F_1( \boldsymbol{\alpha})$ has
different structure regarding $\boldsymbol{\alpha}$ from
$F_2(\boldsymbol{\alpha},\boldsymbol{\beta})$ and
$F_3(\boldsymbol{\alpha},\boldsymbol{\gamma})$.
This may lead to large difficulty in utilizing the common SVM
software package, like LibSVM \cite{Chang01}. To be applied conveniently, we reformulate the benchmark model of Eq. (\ref{Trans1}) for \textit{pTsm}-SVM as follows.

\begin{proposition}
The optimization problem
\begin{eqnarray}\label{Trans4}
&\displaystyle \ \ &\min_{\tau, \boldsymbol{\alpha}\in \mathcal{A}, \theta\geq 0, \boldsymbol{\zeta}\geq \mathbbold{0}_{q^+}, \boldsymbol{\varsigma}\geq \mathbbold{0}_{q^-}}\ -\tau+\boldsymbol{\lambda}^\top \boldsymbol{l}, \\
&\text{s.t.}& \mathcal{L}(\boldsymbol{\alpha})\geq
\tau-\theta, \nonumber\\
& &\alpha_j^+y_j^+[f(\textbf{x}^+_j)-b^*+\boldsymbol{v}_+^\top g^+(\textbf{x}^+_j)+\zeta_j] \geq 0, \forall j,  \nonumber\\
& &\alpha_h^-y_h^-[f(\textbf{x}^-_h)
+b^*-\boldsymbol{v}_-^\top g^-(\textbf{x}^-_h)-\varsigma_h] \geq
0, \forall h \nonumber
\end{eqnarray}
is equivalent to \textit{pTsm}-SVM in Eq. (\ref{Trans1}), where $\alpha_j^+$ and $\alpha_h^-$ are entries in $\boldsymbol{\alpha}$ corresponding to positive and negative class samplings, respectively.
\end{proposition}

\begin{IEEEproof}
Since $\forall j$, $\alpha_j^+\geq 0$ and
$y_j^+=+1$, and $\forall h$, $\alpha_h^-\geq 0$ and $y_h^-=-1$, the
last two constraints in this equation are essentially the same as
the corresponding two constraints emerging in Eq. (\ref{Trans1}). Therefore,
the result is true.
\end{IEEEproof}

Consider all losses in Eq.(\ref{Trans4}) induced by the hinge loss function, then we have the proposition as follows.

\begin{proposition}
The following optimization problem
\begin{equation}\label{Trans5}
\min_{\boldsymbol{\alpha}\in \mathcal{A}}\max_{
\widetilde{\boldsymbol{\beta}}, \widetilde{\boldsymbol{\gamma}}}~~
F_1(
\boldsymbol{\alpha})+\widetilde{F}_2(
\boldsymbol{\alpha},\widetilde{\boldsymbol{\beta}})+\widetilde{F}_3(
\boldsymbol{\alpha},\widetilde{\boldsymbol{\gamma}})
\end{equation}
shares the same solution with \textit{pTsm}-SVM in Eq. (\ref{Trans4}) if
all the losses induced by the slack variables $\theta$,
$\boldsymbol{\zeta}$ and $\boldsymbol{\varsigma}$ obey the rule of
the hinge loss, where
\begin{equation*}\label{waveF2F3}\left\{
\begin{array}{ll}
\widetilde{F}_2(
\boldsymbol{\alpha},\widetilde{\boldsymbol{\beta}})=-
\sum_{j=1}^{q^+}\widetilde{\beta}_j\alpha_j^+[f(\textbf{x}^+_j)-b^*+\boldsymbol{v}_+^\top g^+(\textbf{x}^+_j)],\\
\widetilde{F}_3(
\boldsymbol{\alpha},\widetilde{\boldsymbol{\gamma}})=
\sum_{h=1}^{q^-}\widetilde{\gamma}_h\alpha_h^-[f(\textbf{x}^-_h)
+b^*-\boldsymbol{v}_-^\top g^-(\textbf{x}^-_h)],
\end{array}
\right.
\end{equation*}
$\widetilde{\boldsymbol{\beta}}$ and
$\widetilde{\boldsymbol{\gamma}}$ are the corresponding Lagrangian
multiplier vectors.
\end{proposition}

\begin{IEEEproof}
Eq. (\ref{Trans5}) is the dual form of Eq.
(\ref{Trans4}) under the given conditions, and they thus have the
same solution.
\end{IEEEproof}

\begin{remark}
The decision variables $\widetilde{\boldsymbol{\beta}}$ and
$\widetilde{\boldsymbol{\gamma}}$ in Eq. (\ref{Trans5}) have bounds as $\forall j,~0\leq \widetilde{\beta}_j\leq
\frac{\lambda_2}{\alpha_j^+}$ and $\forall h,~0\leq \widetilde{\gamma}_h\leq \frac{\lambda_3}{\alpha_h^-}$.
\end{remark}

For the convenience of using the common software package,
further rewriting the optimization objective of Eq. (\ref{Trans5}) yields
\begin{equation}\label{Trans6}
\min_{\boldsymbol{\alpha}\in \mathcal{A}}\max_{
\widetilde{\boldsymbol{\beta}}, \widetilde{\boldsymbol{\gamma}}}~~
G_1(
\boldsymbol{\alpha},\widetilde{\boldsymbol{\beta}},\widetilde{\boldsymbol{\gamma}})-G_2(\boldsymbol{\alpha},\widetilde{\boldsymbol{\beta}},\widetilde{\boldsymbol{\gamma}}),
\end{equation}
where
\begin{equation*}\label{G1G2}\left\{
\begin{array}{ll}
G_1(\boldsymbol{\alpha},\widetilde{\boldsymbol{\beta}},\widetilde{\boldsymbol{\gamma}})=F_1(
\boldsymbol{\alpha})-
\sum_{j=1}^{q^+}\widetilde{\beta}_j\alpha_j^+\textbf{z}^\top\textbf{k}(\textbf{x}_j^+)\\
~~~~~~~~~~~~~~~~~~~+\sum_{h=1}^{q^-}\widetilde{\gamma}_h\alpha_h^-\textbf{z}^\top\textbf{k}(\textbf{x}_h^-),\\
G_2(\boldsymbol{\alpha},\widetilde{\boldsymbol{\beta}},\widetilde{\boldsymbol{\gamma}})=\sum_{j=1}^{q^+}\widetilde{\beta}_j\alpha_j^+[b-b^*+\boldsymbol{v}_+^\top g^+(\textbf{x}^+_j)]
\\~~~~~~~~~~~~~~~~~~~-\sum_{h=1}^{q^-}\widetilde{\gamma}_h\alpha_h^-[b
+b^*-\boldsymbol{v}_-^\top g^-(\textbf{x}^-_h)].
\end{array}
\right.
\end{equation*}
Clearly, the degenerated \textit{pTsm}-SVM in Eq. (\ref{Trans6}) has a very similar structure to the
``pure" black-box SVM in Eq. (\ref{SVM4}) with
$G_1(\boldsymbol{\alpha},\widetilde{\boldsymbol{\beta}},\widetilde{\boldsymbol{\gamma}})$ and
$G_2(\boldsymbol{\alpha},\widetilde{\boldsymbol{\beta}},\widetilde{\boldsymbol{\gamma}})$ individually representing
the terms related to the kernel function and to the Lagrangian multipliers. This makes Eq. (\ref{Trans6}) look like a ``standard" SVM so that the corresponding nonlinear QP problem can be easily solved using LibSVM \cite{Chang01}.

{The incorporation mechanism of priors into the black-box soft-margin SVM shown above combines ideas from Mangasarian et. al.\mbox{\cite{Mangasarian08}} and Xu et. al.\mbox{\cite{Xu13}}. Despite this fact, there are still some encouraging novelty in formulating \textit{pTsm}-SVM. Except that the prior information is not ready but needs to be mined from data, the QP problem is handled according to \textbf{Propositions 4} and \textbf{5}, and finally takes the form of Eq. (\mbox{\ref{Trans6}}) which has the same structure as a ``pure" black-box SVM and can thus be solved directly utilizing the existing software packages developed for the standard SVM. More importantly, the incorporation pattern encourages the SVM model to perform classification task following rules, i.e., the mined priors, so the black-box SVM model changes to be interpretable. As a result, the \textit{pTsm}-SVM has the advantage of white-box models as well as of black-box models, i.e., interpretability together with high precision.}

\section{An Algorithm for Mining Data-based Linear Prior Information}
Generally, it is not easy to mine data-based prior information, especially when the feature variables relation contained in $g^+(\textbf{x})$ or $g^-(\textbf{x})$ is nonlinear or the input dimension $n$ is high, even if at $n=3$. For this reason, we only consider linear prior information generated from two feature variables of system. The main thought of mining data-based linear prior information is to map the sampling points in $\mathbb{R}^n$ to the $i$-$j$ ``input feature subspace", in which a linear relation between feature variables $x^{(i)}$ and $x^{(j)}$ is found to be able to separate a class of samplings completely while separate another class of samplings as many as possible. For example, let $\mathscr{L}_{ij}^+$ be a straight line representing the positive class prior information function $g^+(\textbf{x})=\mathbbold{0}_{r^+}$, then it requires that all negative class samplings fall above $\mathscr{L}_{ij}^+$ while positive class samplings fall below $\mathscr{L}_{ij}^+$ as many as possible. The same principle could generate negative class prior information $\mathscr{L}_{ij}^-$. We formulate the mining process as \textbf{Algorithm 1}.

In \textbf{Algorithm 1}, we propose a cardinality maximization problem. The object function $$\text{Card}\left(\{({{x}^{+}}^{(i)},{{x}^{+}}^{(j)})|{{x}^{+}}^{(i)}\cos\phi+{{x}^{+}}^{(j)}\sin\phi+c\leq 0\}\right)$$ represents in the $i$-$j$ input feature space there are as many as possible positive class samples below the learned hyperplane while the constraint ${{x}^{-}}^{(i)}\cos\phi+{{x}^{-}}^{(j)}\sin\phi+c \textgreater 0$ means all negative class samples to be distributed above the hyperplane. Therefore, the solution must classify right as many as possible positive class samples while classify right all negative class samples. In a similar way, the negative class knowledge must classify right as many as possible negative class samples while classify right all positive class samples. Overall the mined linear priors can complement black-box learning algorithms for classification, and can help to solve the original problem.

\begin{figure}[!t]
\begin{center}
\begin{minipage}{\columnwidth}
\begin{small}
\rule{\columnwidth}{1pt} \textbf{Algorithm 1.}{ \textit{Mining Data-based Linear Prior Information}}\\
\rule[1ex]{\columnwidth}{1pt}
\vspace{-1ex} \textbf{Input:} \{$(\textbf{x}_k,y_k)$\}$_{k=1}^N$, $\textbf{x}=(x^{(1)},\cdots,x^{(n)})^\top\in\mathbb{R}^n,
y_k\in \{+1,-1\}$\\
\textbf{Output:} Data-based prior information
\begin{description}
\item[\hspace{-0.5em}1:]\hspace{-0.5em} Data normalization;
\item[\hspace{-0.5em}2:]\hspace{-0.5em} Set $\Gamma^+=\{\textbf{x}_k|y_k=+1,\forall~k\}$ and $\Gamma^-=\{\textbf{x}_k|y_k=-1,\forall~k\}$;
\item[\hspace{-0.5em}3:]\hspace{-0.5em} Find positive class prior information:
\item[\hspace{0em}]\hspace{-1.0em}{\bfseries for} $i=1$ to $n-1:1$ {\bfseries {do}}
\item[\hspace{0em}]\hspace{0em}{\bfseries for} $j=i+1$ to $n:1$ {\bfseries {do}}
\item[\hspace{0em}]\hspace{0.5em}$\circ$ Map all $\textbf{x}_k$ ($k=1,\cdots,N$) to the $i$-$j$ input feature subspace with images denoted by $\text{Img}_{ij}\textbf{x}_k$;
\item[\hspace{0em}]\hspace{0.5em}$\circ$ Solve the following optimization problem
 \begin{eqnarray*}
\hspace{-0.55in} \max_{\phi,c}~~ \text{Card}\left(\{({{x}^{+}}^{(i)},{{x}^{+}}^{(j)})|{{x}^{+}}^{(i)}\cos\phi+{{x}^{+}}^{(j)}\sin\phi+c\leq 0\}\right) \\
\hspace{-3.45in}\text{s.t.} ~~{{x}^{-}}^{(i)}\cos\phi+{{x}^{-}}^{(j)}\sin\phi+c \textgreater 0,~\phi\in[0,2\pi], ~c\in\mathbb{R},~~~~~~~~~\\
\hspace{-0.55in}({{x}^{+}}^{(i)},{{x}^{+}}^{(j)})\in\text{Img}_{ij}\Gamma^+, ~({{x}^{-}}^{(i)},{{x}^{-}}^{(j)})\in\text{Img}_{ij}\Gamma^-.~~~~~~~
\end{eqnarray*}
\item[\hspace{0em}]\hspace{-0.5em}~~where Card($\cdot$) represents the number of elements in set;
 \item[\hspace{0em}]\hspace{-0.5em}$\circ$~Store the optimal results in $\Omega_{(i,j;\hat{\phi}_{ij},\hat{c}_{ij})}$;
\item[\hspace{0em}]\hspace{0em}{\bfseries end for}
\item[\hspace{0em}]\hspace{-1.0em} {\bfseries end for}
\item[\hspace{-0.5em}4:]\hspace{-0.5em} Denote $(\hat{i},\hat{j};\hat{\phi}_{\hat{i}\hat{j}},\hat{c}_{\hat{i}\hat{j}})=\text{arg} \max\limits_{i,j} \Omega_{(i,j;\hat{\phi}_{ij},\hat{c}_{ij})}$;
\item[\hspace{-0.5em}5:]\hspace{-0.5em} Output positive class prior information
\begin{flalign*}
\hspace{-0.45in}\{k|x_k^{(\hat{i})}\cos\hat{\phi}_{\hat{i}\hat{j}}+x^{(\hat{j})}_k\sin\hat{\phi}_{\hat{i}\hat{j}}+\hat{c}_{\hat{i}\hat{j}}\leq 0\Longrightarrow y_k=+1\};
\end{flalign*}
\item[\hspace{-0.5em}6:]\hspace{-0.5em} The same way produces negative class prior information.
\end{description}
\rule[0ex]{\columnwidth}{1pt}\hspace{2mm}
\end{small}
\end{minipage}
\end{center}
\end{figure}

Seemingly, the optimization in \textbf{Algorithm 1} is a NP hard nonconvex/nonlinear/nonsmooth optimization problem. However, in fact the hyperplane learned by optimizing $\phi$ and $c$ is simply a normalized hyperplane (that is a unit-norm hyperplane). Theoretically, the hyperplane optimization could be rewritten as a linear program with respect to the normal vector to the line instead of $\phi$ and $c$. If doing so, the cardinality maximization problem would become difficult to be solved with respect to the normal vector. The usage of the angle $\phi$ instead can avoid this issue. In practical solution procedure, the angel range $[0,2\pi]$ is discretized into $63$ points with step by $0.1$, and the hyperplane is rotated along every discretized angle to produce the result. We admit that this result is not the optimal one, but a relaxation solution. It naturally can produce better solution by making more thinning discretization on the angle.

Also, it should be noted in \textbf{Algorithm 1} that there needs to solve $\frac{n(n-1)}{2}$ optimization problems to find a piece of prior information. The computational complexity of the algorithm is thus $O(63\times N\times \frac{n(n-1)}{2})$. It is obvious that the algorithm scalability will increase as the the dimension of data, $n$, becomes larger. However, compared with the SVM model solution process, \textbf{Algorithm 1} will not lead to scalability disaster even if for high-dimensional data, since the former has the computational complexity $O(n\times N^2)$ \mbox{\cite{Chang01}} while the size of the data set, $N$, is usually far larger than $n$. This means the algorithm can be generalized to handle general classification problems such as text classification and image classification where high dimensional features will be encountered. We exhibit this algorithm through the following example and also report the running time.

\textit{Example 1}. \textbf{Algorithm 1} is applied to Liver disorders\footnote{http://www.csie.ntu.edu.tw/$\sim$cjlin/libsvmtools/datasets}, a public data set, to exhibit effectiveness, in which there are $6$ feature variables $x^{(1)},\cdots,x^{(6)}$, and $345$ recordings with $200$ positive class points while $145$ negative class points. We select $70\%$ sampling points randomly and feed them into the mining algorithm. Table \ref{Example1} reports the mined positive class linear prior information along every projected two dimensional plane. It is clear that the straight line $$\mathscr{L}_{(1,4;\hat{\phi}_{14},\hat{c}_{14})}^+:0.6347x^{(1)}-0.7728x^{(4)}-0.0156=0$$ in the $1$-$4$ input feature subspace is the expected one. The running time is about $0.30$s, which is very small. In the same way, we can obtain negative class prior information in the $3$-$5$ input feature subspace to be
$$\mathscr{L}_{(3,5;\hat{\phi}_{35},\hat{c}_{35})}^-:-0.1288x^{(3)}+0.9917x^{(5)}+0.0025=0.$$

\begin{remark}
The linear prior information actually represents a boundary in a plane related to two feature variables of system. Every boundary serves for the largest degree of separation between two classes in the plane constructed by two particular variables. For ``positive"/``negative" boundary, all negative/positive class samplings fall above it while positive/negative class samplings fall below it as many as possible.
\end{remark}

\begin{table}
\renewcommand\arraystretch{1.5}
\centering \caption{Mined Positive Class Linear Priors for Liver Disorders Data Set along Two Dimensional Input Feature Spaces}\label{Example1}
\begin{tabular}{|m{37pt}<{\centering}| m{139pt}<{\centering}| m{40pt}<{\centering}|}
\hline \tabincell{c}{$i$-$j$ } & \tabincell{c}{$\mathscr{L}_{(i,j;\hat{\phi}_{ij},\hat{c}_{ij})}^+$ } & \tabincell{c}{$\Omega_{(i,j;\hat{\phi}_{ij},\hat{c}_{ij})}$}
\\
\hline
\tabincell{c}{$1$-$2$}&
\tabincell{l}{$-0.9422x^{(1)}+0.3350x^{(2)}+0.7347\leq 0$} & \tabincell{c}{$5$} \\
\hline
\tabincell{c}{$1$-$3$}&
\tabincell{l}{$0.1700x^{(1)}+0.9854x^{(3)}-0.1651\leq 0 $} & \tabincell{c}{$10$} \\
\hline
\tabincell{c}{$1$-$4$}&
\tabincell{l}{$0.6347x^{(1)}-0.7728x^{(4)}-0.0156\leq 0 $} & \tabincell{c}{$\textbf{13}$} \\
\hline
\tabincell{c}{$1$-$5$}&
\tabincell{l}{$0.5403x^{(1)}+0.8415x^{(5)}-0.2648\leq 0 $} & \tabincell{c}{$4$} \\
\hline
\tabincell{c}{$1$-$6$}&
\tabincell{l}{$-0.9900x^{(1)}+0.1411x^{(6)}+0.8689\leq 0$} & \tabincell{c}{$6$} \\
\hline
\tabincell{c}{$2$-$3$}&
\tabincell{l}{$-0.7259x^{(2)}-0.6878x^{(3)}+0.7686\leq 0 $} & \tabincell{c}{$5$} \\
\hline
\tabincell{c}{$2$-$4$}&
\tabincell{l}{$-0.7910x^{(2)}-0.6119x^{(4)}+0.8974\leq 0 $} & \tabincell{c}{$5$} \\
\hline
\tabincell{c}{$2$-$5$}&
\tabincell{l}{$-0.8968x^{(2)}-0.4425x^{(5)}+0.8473\leq 0 $} & \tabincell{c}{$3$} \\
\hline
\tabincell{c}{$2$-$6$}&
\tabincell{l}{$-0.8011x^{(2)}+0.5985x^{(6)}+0.6779\leq 0 $} & \tabincell{c}{$2$} \\
\hline
\tabincell{c}{$3$-$4$}&
\tabincell{l}{$0.9950x^{(3)}+0.0998x^{(4)}-0.0634\leq 0 $} & \tabincell{c}{$6$} \\
\hline
\tabincell{c}{$3$-$5$}&
\tabincell{l}{$0.9801x^{(3)}+0.1987x^{(5)}-0.0510\leq 0 $} & \tabincell{c}{$4$} \\
\hline
\tabincell{c}{$3$-$6$}&
\tabincell{l}{$-0.5048x^{(3)}+0.8632x^{(6)}+0.1437\leq 0 $} & \tabincell{c}{$6$} \\
\hline
\tabincell{c}{$4$-$5$}&
\tabincell{l}{$-0.8011x^{(4)}+0.5985x^{(5)}+0.3473\leq 0 $} & \tabincell{c}{$7$} \\
\hline
\tabincell{c}{$4$-$6$}&
\tabincell{l}{$-0.5885x^{(4)}+0.8085x^{(6)}+0.2136\leq 0 $} & \tabincell{c}{$9$} \\
\hline
\tabincell{c}{$5$-$6$}&
\tabincell{l}{$-0.9422x^{(5)}+0.3350x^{(6)}+0.4111\leq 0 $} & \tabincell{c}{$5$} \\
\hline
\end{tabular}
\end{table}

\begin{remark}
It is possible to generate multi sets of positive priors or of negative ones for an objective system. As an example of positive prior information, this takes place when there are more than two projected two dimensional planes in which the positive ``boundaries" frame the same number positive samplings below them. In this case, the multi sets of priors can be applied at the same time without worrying that they are conflicting since they come from different planes.
\end{remark}

\begin{remark}
Compared with the high dimensional nonlinear model of system, the linear prior information only including two feature inputs is quite simple. However, it provides an available way to mine the knowledge hidden in data. An interesting and challenging issue for future research is to include more inputs or to use nonlinear function with respect to inputs to model prior information. It may be relatively simple to construct a separating plane as prior information that is captured by a linear function with respect to three feature inputs.
\end{remark}

\section{Numerical Experiments}
In this section, the degenerated \textit{pTsm}-SVM in Eq. (\ref{Trans6}) is used to model $8$ benchmark data sets (all available at the same website as given in \textit{Example 1}): Australian, Breast cancer, Diabetes, German, Heart, Ionosphere, Liver disorders and Sonar. Some basic information about them, like size and number of feature variables, is exhibited in Table \ref{UCI1}. For these $8$ benchmark examples, two-class classification problems are addressed. The priors for every data set are linear and acquired through the proposed mining algorithm in Section V. For the kernel function in Eq. (\ref{Trans6}), the Gaussian radial basis kernel defined by
\begin{eqnarray}\label{Gaussiankernel}
k(\textbf{x}_i,\textbf{x}_j)=\text{exp}(-\|\textbf{x}_i-\textbf{x}_j\|^2/\sigma^2)
\end{eqnarray}
is used. For every data set, the recordings are segmented into two groups, one group including $70\%$ samplings as training set, and the other group including the remaining $30\%$ samplings as testing set. The training set serves for generating linear priors and learning parameters while the testing set works for evaluating performance of the degenerated \textit{pTsm}-SVM in Eq. (\ref{Trans6}). The parameters training is made through grid search together with
five-fold crossing validation for the purpose of reducing over-fitting phenomenon. {All experiments are carried out in Matlab7.0 environment running on a desktop PC with a 1.80GHz AMD Athlon (tm) Processor and a 4.00 GB memory.}

\begin{table}
\centering \caption{Benchmark Data Sets Information} \label{UCI1}
\begin{tabular}{|l|c|c|}
\hline \tabincell{c}{Name of \\Data Set} & \tabincell{c}{Size of Recordings\\ (positive/negative)} &
\tabincell{c}{Number of \\Feature Variables}\\
\hline
\tabincell{c}{Australian}& \tabincell{c}{690 (307/383)}& \tabincell{c}{14}\\
\hline
\tabincell{c}{Breast cancer}& \tabincell{c}{683 (239/444)}& \tabincell{c}{10}\\
\hline
\tabincell{c}{Diabetes}& \tabincell{c}{768 (268/500)}& \tabincell{c}{8}\\
\hline
\tabincell{c}{German}& \tabincell{c}{1000 (300/700)}& \tabincell{c}{24}\\
\hline
\tabincell{c}{Heart}& \tabincell{c}{270 (120/150)}& \tabincell{c}{13}\\
\hline
\tabincell{c}{Ionosphere}& \tabincell{c}{351 (225/126)}& \tabincell{c}{34}\\
\hline
\tabincell{c}{Liver disorders}& \tabincell{c}{345 (200/145)}& \tabincell{c}{6}\\
\hline
\tabincell{c}{Sonar}& \tabincell{c}{208 (97/111)}& \tabincell{c}{60}\\
\hline
\end{tabular}
\end{table}

{
\begin{table*}
\renewcommand\arraystretch{1.5}
\centering \caption{Acquired Prior Information and Classification
Results of Benchmark Data Sets} \label{UCI}
\begin{tabular}{|l|l|c|c|c|c|}
\hline \tabincell{c}{Name of \\Data Set} & \tabincell{c}{~~~~~~~~~~~~~~~~~~~~~~~~~Linear Prior Information} &\tabincell{c}{\textbf{Algorithm 1} \\time cost (s)}& \tabincell{c}{ATA$^\dag$ (\%)\\with Priors}&
\tabincell{c}{ATA (\%)\\without Priors}& \tabincell{c}{{p-value}\\{of the t-test}} \\
\hline
\tabincell{c}{Australian}&
\tabincell{l}{$g^+(\textbf{x})=-0.3037x^{(5)}-0.9516x^{(14)}+0.3085\leq 0 \Longrightarrow y=+1$
\\$g^-(\textbf{x})=0.7087x^{(8)}-0.7055x^{(10)}+0.0176\leq 0 \Longrightarrow y=-1$} & \tabincell{c}{$0.54$}&\tabincell{c}{$85.22\pm 2.70$ } & \tabincell{c}{$82.54\pm 6.40$}&$0.03$\\
\hline
\tabincell{c}{Breast cancer}&
\tabincell{l}{$g^+(\textbf{x})=0.8855x^{(1)}-0.4646x^{(9)}+0.2890\leq 0 \Longrightarrow y=+1$
\\$g^-(\textbf{x})=0.9211x^{(1)}+0.3894x^{(8)}-0.0746\leq 0 \Longrightarrow y=-1$} &\tabincell{c}{$0.30$} &\tabincell{c}{$97.34\pm 0.53$ } & \tabincell{c}{$96.85\pm 0.58$}&$0.00$\\
\hline
\tabincell{c}{Diabetes}&
\tabincell{l}{$g^+(\textbf{x})=-0.9900x^{(2)}+0.1411x^{(7)}+0.9467\leq 0 \Longrightarrow y=+1$
\\$g^-(\textbf{x})=0.3624x^{(2)}+0.9320x^{(8)}-0.2004\leq 0 \Longrightarrow y=-1$} &\tabincell{c}{$0.30$} &\tabincell{c}{$76.67\pm 1.66$ } & \tabincell{c}{$75.92\pm 1.26$}&$0.04$\\
\hline
\tabincell{c}{German}&
\tabincell{l}{$g^+(\textbf{x})=0.6347x^{(2)}-0.7728x^{(4)}+0.3679\leq 0 \Longrightarrow y=+1$
\\$g^-(\textbf{x})=0.0875x^{(4)}-0.9962x^{(17)}+0.9812\leq 0 \Longrightarrow y=-1$} & \tabincell{c}{$0.89$} &\tabincell{c}{$75.83\pm 2.44$} & \tabincell{c}{$74.83\pm 2.23$}& $0.05$\\
\hline
\tabincell{c}{Heart}&
\tabincell{l}{$g^+(\textbf{x})=-0.3739x^{(10)}+0.9275x^{(8)}-0.0547\leq 0 \Longrightarrow y=+1$
\\$g^-(\textbf{x})=0.7756x^{(5)}-0.6317x^{(8)}+0.4139\leq 0 \Longrightarrow y=-1$} & \tabincell{c}{$0.31$}&\tabincell{c}{$84.25\pm 3.74$} & \tabincell{c}{$83.50\pm 4.78$}&  $0.26$\\
\hline
\tabincell{c}{Ionosphere}&
\tabincell{l}{$g^+(\textbf{x})=-0.6663x^{(5)}+0.7457x^{(27)}+0.3465\leq 0 \Longrightarrow y=+1$
\\$g^-(\textbf{x})=x^{(5)}-0.5210\leq 0 \Longrightarrow y=-1$} &\tabincell{c}{$3.40$} &\tabincell{c}{$93.96\pm 2.23$ } & \tabincell{c}{$93.49\pm 1.69$}&$0.21$\\
\hline
\tabincell{c}{Liver disorders}&
\tabincell{l}{$g^+(\textbf{x})=0.6347x^{(1)}-0.7728x^{(4)}-0.0156\leq 0 \Longrightarrow y=+1$
\\$g^-(\textbf{x})=-0.1288x^{(3)}+0.9917x^{(5)}+0.0025\leq 0 \Longrightarrow y=-1$} & \tabincell{c}{$0.30$} &\tabincell{c}{$73.10\pm 3.70$ } & \tabincell{c}{$70.80\pm 3.52$}&$0.02$\\
\hline
\tabincell{c}{Sonar}&
\tabincell{l}{$g^+(\textbf{x})=0.9553x^{(13)}+0.2955x^{(20)}-0.2426\leq 0 \Longrightarrow y=+1$
\\$g^-(\textbf{x})=-0.5748x^{(11)}-0.8183x^{(27)}+0.9725\leq 0 \Longrightarrow y=-1$} & \tabincell{c}{$3.61$}& \tabincell{c}{$85.72\pm 3.59$ } & \tabincell{c}{$83.02\pm 4.55$}&$0.04$\\
\hline
 \multicolumn{3}{l}{\hspace{-2mm}\scriptsize{$^\dag$ represents Average Testing Accuracy.}}
\end{tabular}
\end{table*}
}

The experiments begin with normalizing all input features in the training recordings to the
range $[0,1]$. Then we apply the proposed mining algorithm to acquire the linear priors for every data set. Shown in Table \ref{UCI} are the results, where the running time for every data set is also reported. As can be seen, the longest time is $3.61$s for Sonar data set which further indicates that \textbf{Algorithm 1} has large potential to be generalized to handle general classification problems. Note that the inclusion of priors may bring a great deal of equations into the optimization model, especially when the meshes of points are high. It is possible that these equations are either redundant or causing overfitting in machine learning. To circumvent this issue, we only consider those prior constraints on data points. As an example, there are $13$ equations, i.e., $q^+=13$, as positive class priors included for Liver data set. After substituting these priors into Eq. (\ref{Trans6}), respectively, there are $(\nu,\sigma,\widetilde{\boldsymbol{\beta}}, \widetilde{\boldsymbol{\gamma}})$ left to be trained. Here, we use grid search to find these optimal parameters. For the first parameter $\nu$, its physical meaning implies it should not be too high, i.e., requesting enough good training, but to avoid getting in over-fitting, it may not be too low, i.e., training not allowed to be too good. We thus set a $10$-point uniform discretization in $[\nu_{\text{min}},\nu_{\text{max}}]$ as the searching range for finding it with $\nu_{\text{min}}=0.1$ and $\nu_{\text{max}}$ depending on the specific example, calculated by $\nu_{\text{max}}=\frac{2\text{min}(N^+,N^-)}{N}$ \cite{Hsuen05} where $N^+$ and $N^-$ represent the number of positive samplings and negative ones, respectively. As for the other three parameters, the searching ranges are $\{2^{-3}, 2^{-2}, \cdots, 2^{5}, 2^{6}\}$ for $\sigma$, and $\frac{1}{N}$ times of some points in $[0,1]$ for the components of $\widetilde{\boldsymbol{\beta}}$ and $\widetilde{\boldsymbol{\gamma}}$. We consult the penalty factor of slack variables $\xi_i~(i=1,\cdots,N)$ in Eq. (\ref{MSVM2}) to set the searching range for $\widetilde{\boldsymbol{\beta}}$ and $\widetilde{\boldsymbol{\gamma}}$ like so.

After finishing learning these parameters through grid search and five-fold crossing validation, we further use the testing samplings to evaluate the performance of the degenerated \textit{pTsm}-SVM. To make the results more convincing, we have carried out $10$ times random experiments for every example, and calculated their average testing accuracy values and the corresponding standard deviations, as shown in Table \ref{UCI}. Here, the testing accuracy is defined by the ratio of the number of testing samplings to be classified right to the total number of testing samples. In each experiment, $70\%$ samplings are selected randomly as the training set while the remaining $30\%$ samplings are set as the testing set. The average testing accuracy (ATA) reported in Table \ref{UCI} suggests that the degenerated \textit{pTsm}-SVM basically can perform the $2$-class classification task for these $8$ benchmark examples well, high ATA but low standard deviations.

To further exhibit the performance of the degenerated \textit{pTsm}-SVM, we have made some parallel experiments on these data sets using the soft-margin SVM model, i.e., without priors incorporated. The testing results are also reported in Table \ref{UCI}. {It is clear that the degenerated \textit{pTsm}-SVM outperforms the soft-margin SVM for all benchmark data sets according to the ATA. Moreover, the stability of the ATA, characterized by the standard deviation, for some data sets is also strengthened after incorporating priors, such as for Australian, Heart and Sonar data sets where much smaller standard deviations emerge. For Breast cancer data set, the ATA stability basically keeps unchanged; but for Diabetes, German, Ionosphere and Liver disorders data sets, the ATA stability changes a little weaker when priors are integrated. The conflict phenomena reflected by the ATA and standard deviation for the last four mentioned data sets mean it difficult to say the inclusion of linear priors playing a positive role on improving precision for them. Even though for the Australian-like data sets, it is still difficult to say that the degenerated \textit{pTsm}-SVM must have higher precision than the soft-margin SVM, as the extreme case $(85.22-2.70)$ in the former model is apparently lower than the extreme case $(82.54+6.40)$ in the latter one. To achieve rigorous comparisons, we make a paired Student's t-test on the classification results produced by the used two kinds of models. The t-test results (p-values) are also reported in Table \mbox{\ref{UCI}}, where the p-value represents the probability in which the ATA of the soft-margin SVM model is no less than the ATA of the degenerated \textit{pTsm}-SVM with the significant level of $0.05$. Clearly, except the data sets of Heart and Ionosphere, the other $6$ data sets exhibit that the integration of the mined priors can improve the precision of the soft-margin SVM model in very high probability, i.e., in these $6$ data sets the degenerated \textit{pTsm}-SVM is statistically significantly better than the soft-margin SVM. Therefore, from these $6$ examples, it might suggest that the degenerated \textit{pTsm}-SVM, on the one hand, has higher interpretability than the soft-margin SVM (the structure and the solving algorithm of the black-box soft-margin SVM are changed due to incorporation of the mined linear priors, and moreover, the mined linear priors are highly related to the background of the corresponding data set); on the other hand, the former has larger possibility to make right classifications than the latter. As for the data sets of Heart and Ionosphere, although the degenerated \textit{pTsm}-SVM is more interpretable than the soft-margin SVM, it cannot outperform the latter statistically significantly in precision, the p-values only being $26\%$ and $21\%$, respectively. The possible reason for these two high p-values may be that the linear priors mined for these two data sets are not so good, even not true, which sometimes play a constructive role while sometimes play a negative role. A solution to overcome this issue may be either to integrate other priors with respect to more features instead or to relax the current ones so that a little less positive/negative points are included below the ``positive"/``negative" boundaries. Despite the possibility of slight loss in precision for these two data sets, the degenerated \textit{pTsm}-SVM is still a good alternative due to its interpretability and larger potential of practical applications. }

To exhibit the reliability of the degenerated \textit{pTsm}-SVM in precision, we present the experimental results on some of those benchmark examples produced by competing state-of-the-arts in Table \ref{UCI2gao}. The model in \mbox{\cite{Xu13}} is a soft-margin multiple kernels SVM while the model in \mbox{\cite{Liu13}} is multiple kernels SVM integrating radius information. These two kinds of models were asserted to be able to outperform other similar models, like MKL, $\ell_p$MKL, etc. As can be seen from Table \ref{UCI2gao}, for the degenerated $\textit{pTsm}$-SVM and the listed two models each has its own merits, either in ATA or in standard deviation. However, it needs to mention that it is not quite fair to compare the current classification results with those generated by Soft-margin MKL or $\ell_2\text{trSt}$MKL, since the testing samplings are not the same during every random experiment. We list them here not for solid comparisons but only for reference.

\begin{table}
\centering \caption{Classification Results on the Benchmark Data Sets with Other Models} \label{UCI2gao}
\begin{tabular}{|l|c|c|c|}
\hline \tabincell{c}{Name of \\Data Set} & \tabincell{c}{degenerated\\$\textit{pTsm}$-SVM} &
\tabincell{c}{Soft-margin \\MKL \cite{Xu13}$^\S$}& \tabincell{c}{$\ell_2\text{trSt}$MKL \\\cite{Liu13}$^\dag$}\\
\hline
\tabincell{c}{Australian}& \tabincell{c}{$85.22\pm 2.70$}& \tabincell{c}{$86.23\pm 1.94$}& \tabincell{c}{$\varnothing$}\\
\hline
\tabincell{c}{Diabetes}& \tabincell{c}{$76.67\pm 1.66$}& \tabincell{c}{$76.35\pm 2.79$}& \tabincell{c}{$\varnothing$}\\
\hline
\tabincell{c}{German}& \tabincell{c}{$75.83\pm 2.44$}& \tabincell{c}{$\varnothing$}& \tabincell{c}{$74.40\pm 1.00$}\\
\hline
\tabincell{c}{Heart}& \tabincell{c}{$84.25\pm 3.74$}& \tabincell{c}{$81.60\pm 4.21$}& \tabincell{c}{$85.10\pm 1.40$}\\
\hline
\tabincell{c}{Ionosphere}& \tabincell{c}{$93.96\pm 2.23$}& \tabincell{c}{$91.33\pm 2.82$}& \tabincell{c}{$94.70\pm 1.10$}\\
\hline
\tabincell{c}{Liver disorders}& \tabincell{c}{$73.10\pm 3.70$}& \tabincell{c}{$\varnothing$}& \tabincell{c}{$66.20\pm 2.40$}\\
\hline
\tabincell{c}{Sonar}& \tabincell{c}{$85.72\pm 3.59$}& \tabincell{c}{$\varnothing$}& \tabincell{c}{$83.30\pm 2.60$}\\
\hline
\multicolumn{3}{l}{\hspace{-2mm}\scriptsize{$^\S$
ATA over 10 times random experiments;}}\\ \multicolumn{3}{l}{\hspace{-2mm}\scriptsize{$^\dag$
ATA over 30 times random experiments.}}
\end{tabular}
\end{table}

\begin{table}
\renewcommand\arraystretch{1.5}
\centering \caption{{Average Processing Time
for the Benchmark Data Sets}} \label{UCIgao}
\begin{tabular}{|l|c|c|}
\hline \tabincell{c}{Name of \\Data Set}&
\tabincell{c}{APT$^\star$ (s)\\with Priors}&
\tabincell{c}{APT (s)\\without Priors}\\
\hline
\tabincell{c}{Australian}&
\tabincell{c}{$728.9$} & \tabincell{c}{$28.4$}\\
\hline
\tabincell{c}{Breast cancer}&
\tabincell{c}{$220.8$} & \tabincell{c}{$15.5$}\\
\hline
\tabincell{c}{Diabetes}&
\tabincell{c}{$860.8$} & \tabincell{c}{$21.3$}\\
\hline
\tabincell{c}{German}&
\tabincell{c}{$1708.5$} & \tabincell{c}{$70.6$}\\
\hline
\tabincell{c}{Heart}&
\tabincell{c}{$15.1$} & \tabincell{c}{$3.9$}\\
\hline
\tabincell{c}{Ionosphere}&
\tabincell{c}{$927.7$} & \tabincell{c}{$6.8$}\\
\hline
\tabincell{c}{Liver disorders}&
\tabincell{c}{$116.6$} & \tabincell{c}{$4.7$}\\
\hline
\tabincell{c}{Sonar}&
\tabincell{c}{$58.7$} & \tabincell{c}{$4.2$}\\
\hline
 \multicolumn{3}{l}{\hspace{-2mm}\scriptsize{$^\star$ APT represents Average Processing Time.}}
\end{tabular}
\end{table}

{It should be noted that the incorporation of priors into the soft-margin SVM may result in the processing time for the degenerated $\textit{pTsm}$-SVM to increase greatly, including priors mining and soft-margin SVM training time. There are two additional parameter vectors $\widetilde{\boldsymbol{\beta}}$ and $\widetilde{\boldsymbol{\gamma}}$ in the degenerated $\textit{pTsm}$-SVM that need to be trained. Table \mbox{\ref{UCIgao}} reports the average processing time (APT) of the model with and without priors incorporation for the $8$ benchmark data sets. As expected just now, the APT for the degenerated $\textit{pTsm}$-SVM is far beyond that for the soft-margin SVM, especially in the case of the Ionosphere data set, $927.7$s vs. $6.8$s. This terrible phenomenon seems to suggest the degenerated $\textit{pTsm}$-SVM is much poorer than the soft-margin SVM. Actually, as far as these benchmark data sets are concerned, the processing time of the model may be not quite important, such as for the Breast cancer/Heart data sets, the most important should be to diagnose right the cancer/heart disease and also to give the etiology, but not the diagnostic time. Therefore, the reported APT of every benchmark data set with priors in Table \mbox{\ref{UCIgao}}, despite being very high, is still acceptable from the viewpoint of practical applications. Certainly, there are rather large rooms to reduce the processing time of the degenerated $\textit{pTsm}$-SVM so that this model looks also efficient. Since the additional processing time mainly comes from training $\widetilde{\boldsymbol{\beta}}$ and $\widetilde{\boldsymbol{\gamma}}$ in Eq. \mbox{(\ref{Trans6})}, a solution of raising efficiency of the model is to derive their optimal values from theoretical analysis instead of grid search. We put this issue as one of the main points of future research.    }

{In summary, from the viewpoint of model precision, the degenerated $\textit{pTsm}$-SVM is not sure to be significantly superior to some competing state-of-the-arts, and even a simple multi-layer neural network might produce better accuracy in some of the benchmark data sets. However, the largest advantage for the proposed model is that it has interpretability while the others are ``black". In the current modeling framework, there are practical domains knowledge, i.e., the mined linear priors, for every data set before modeling them, which can be thought as known information about the modeling object. The incorporation of these priors into the black-box model, soft-margin SVM, has a large effect on the model structure and the solving algorithm, which results in the interpretability enhancement of the black-box soft-margin SVM. Frankly speaking, it is not our original intension to expect the degenerated $\textit{pTsm}$-SVM better than the competing state-of-the-arts in precision when used for the benchmark data sets. The main contribution of this paper is to provide a way for adding interpretability of black-box models, and the models performance comparison should be made between the interpretable model and the corresponding black-box model. It is interesting to observe the effect if the mined linear priors, maybe in the form of logical implications as given in Eqs. (\mbox{\ref{pi1}}) and (\mbox{\ref{pi2}}), are incorporated into other black-box models, such as neural network, multiple kernels learning\mbox{\cite{Hu09,Liu13}}, etc. To ensure the model performance better, it naturally needs to mine priors as accurate as possible from the data sets, which constitutes one of our main concerns in the future research. In addition, it is very time-consuming for the degenerated $\textit{pTsm}$-SVM to be trained, including priors mining plus SVM training. As an example of the Heart data set, the average processing time is $15.1$s for the degenerated $\textit{pTsm}$-SVM while $3.9$s for the soft-margin SVM. It is possible to avoid this point by achieving the optimal parameters related to the priors, like $\widetilde{\boldsymbol{\beta}}$ and $\widetilde{\boldsymbol{\gamma}}$ in Eq. (\mbox{\ref{Trans6}}), based on theoretical analysis. The effort towards this target is on the way.}

\section{Conclusions and Points of Future Research}
This paper has presented a theoretical and experimental study on the incorporation of data-based prior information into black-box soft-margin SVM model for interpretability and precision enhancement. The main contribution includes: i) propose a concise and practical algorithm to mine linear prior information from data set; ii) for the soft-margin SVM with priors incorporated, develop an equivalent model that seems to have the same structure as ``pure" black-box SVM so that the common software packages can be directly utilized to find solutions.

{Despite the good performance exhibited by the degenerated \textit{pTsm}-SVM, there are still great rooms for the model to be improved. The most urgent task maybe focuses on mining priors as reliable as possible from the data set, such as including more inputs to model linear priors or directly developing nonlinear priors.
Moreover, in the case of the linear priors, the formulation of the optimization problem should consider the entire regions of feature space defined by the prior knowledge constraints, but no longer depends on $\textbf{x}^+$ and $\textbf{x}^-$ as individual data points or meshes of points. After all, the current improvements in data sets are very small when prior knowledge is added while a lot of previous results with prior knowledge report significant accuracy improvements \cite{Mangasarian08,FMS03} of anywhere from $15\%$ to $50\%$. In addition, more concerns should be thrown towards reducing the training time of the interpretable model. The current framework is quite time-consuming since the incorporation of priors would introduce more parameters needed to be studied. The next effort is to estimate the theoretical optimal values of these parameters so that the interpretable model has the same number of parameters to be trained with the corresponding black-box model. Certainly, it is interesting to achieve the theoretical support that the degenerated \textit{pTsm}-SVM is able to converge the true one, i.e., to the corresponding Bayes model. The estimation of the convergence rate is also an important investigation point.}

Finally, it should be pointed out that the prior information learned is irrelevant with SVM. Even if the problem described in Eq. (\ref{dpi}) is derived by imitating the mechanism of SVM, it is actually not specialized for improving the performance of SVM. Those prior information corrected from Eq. (\ref{dpi}) can
also be used in other machine learning algorithms such as naive Bayes or deep neural network. Therefore, it is interesting to investigate the incorporation of the mined priors into other machine learning algorithms for interpretability enhancement.

\ifCLASSOPTIONcompsoc
\else
\fi


\ifCLASSOPTIONcaptionsoff
  \newpage
\fi



%

\begin{IEEEbiography}[{\includegraphics[width=1.3in,height=1.6in,clip,keepaspectratio]{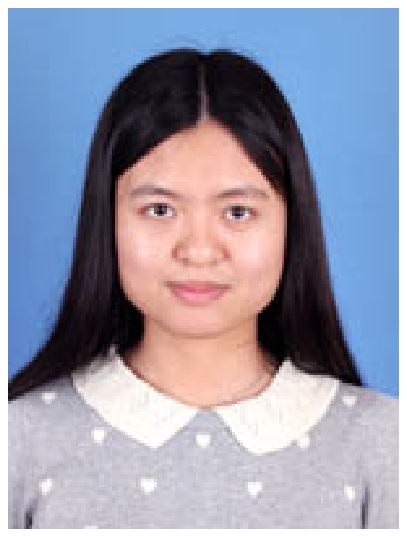}}]
{Shaohan Chen} received the B.S. degrees in Mathematics and Applied Mathematics from Jimei University, China, in 2014.
She is currently working towards the Ph.D. degree in operational research and cybernetics at Zhejiang University.

Her research interests are in the areas of Machine learning and Transparency of black-box modeling techniques.
\end{IEEEbiography}

\begin{IEEEbiography}[{\includegraphics[width=1.3in,height=1.55in,clip,keepaspectratio]{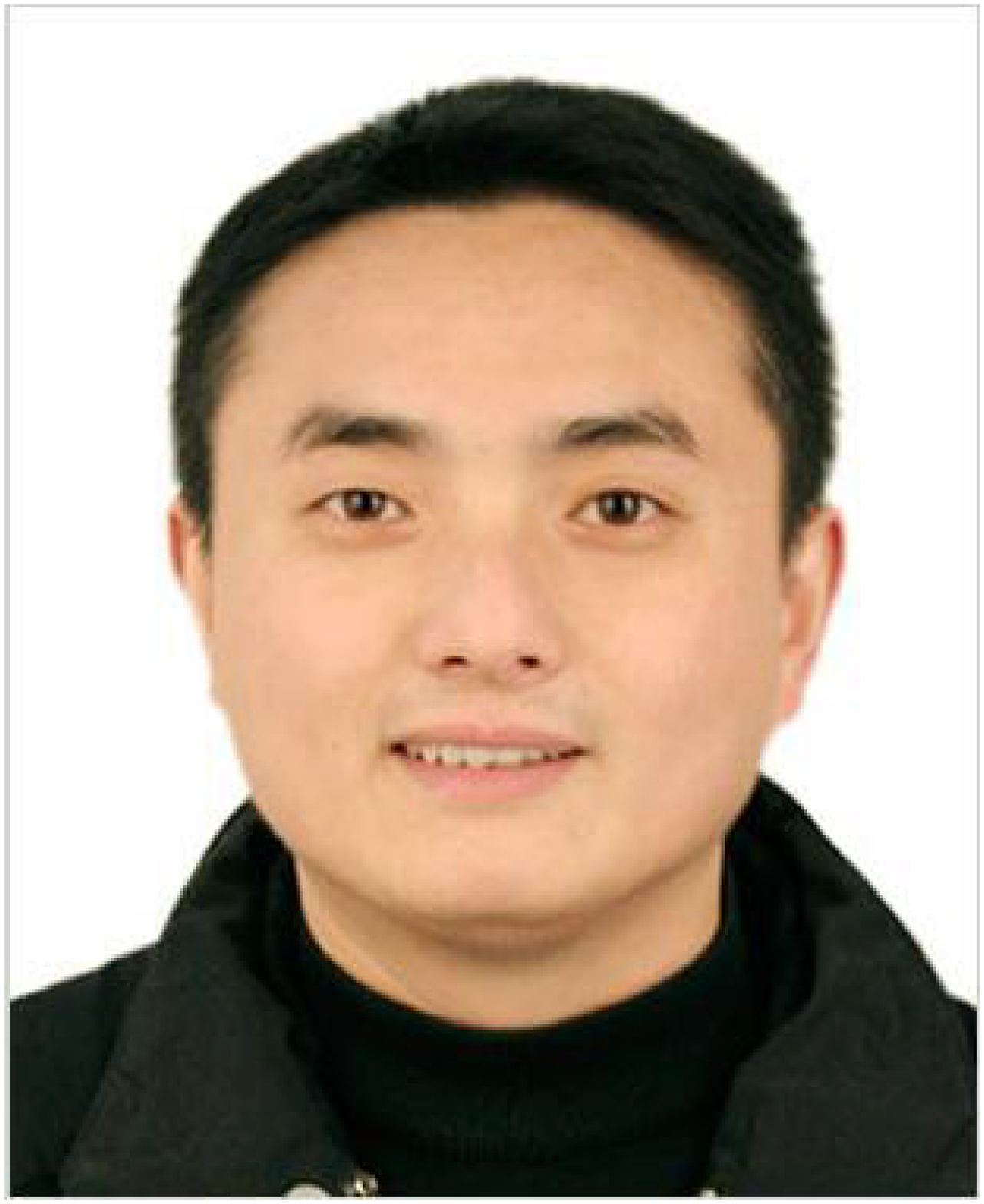}}]
{Chuanhou Gao}(M'09 SM'12) received the B.Sc. degrees in Chemical Engineering from Zhejiang
University of Technology, China, in 1998, and the Ph.D. degrees in
Operational Research and Cybernetics from Zhejiang University,
China, in 2004. From June 2004 until May 2006, he was a Postdoctor
in the Department of Control Science and Engineering at Zhejiang
University.

Since June 2006, he has joined the Department of Mathematics at
Zhejiang University, where he is currently a Professor. He was a
visiting scholar at Carnegie Mellon University from Oct. 2011 to
Oct. 2012. His research interests are in the areas of data-driven
modeling, control and optimization, chemical reaction network theory
and thermodynamic process control. He is a guest editor of IEEE Transactions on
Industrial Informatics, ISIJ International and Journal of Applied
Mathematics, and an editor of Metallurgical Industry Automation from
May 2013.
\end{IEEEbiography}

\begin{IEEEbiography}[{\includegraphics[width=1.25in,height=1.5in,clip,keepaspectratio]{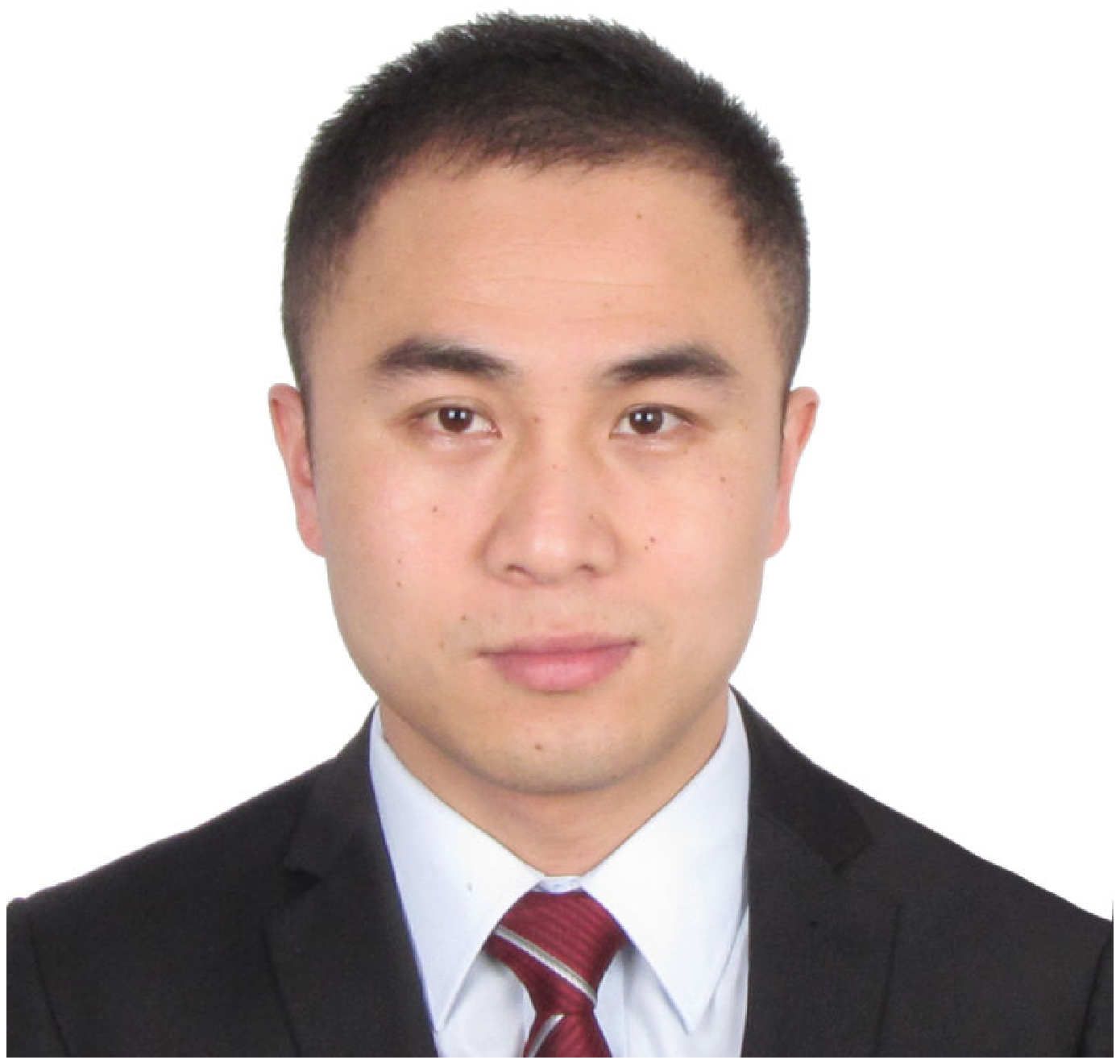}}]
{Ping Zhang} received the B.S. degrees in applied information and computation science from China Ji Liang Univercity, China, in 2012. And the M.S. degrees in operational research cybernetics from Zhejiang University, China, in 2014.

Since July 2014, he has joined the ZTE Co. , Ltd. , Shanghai, China. His research interests are in the areas of Machine learning and Transparency of black-box modeling techniques.
\end{IEEEbiography}
%
%
%




\end{document}